\documentclass[conference]{IEEEtran}
\IEEEoverridecommandlockouts
\usepackage{cite}
\usepackage{amsmath,amssymb,amsfonts}
\usepackage{algorithmic}
\usepackage{textcomp}
\usepackage[pdftex]{graphicx}
\usepackage{pgfplots}
\usepackage{color}
\usepackage{subcaption}
\usepackage{comment}
\usepackage{graphicx}       
\usepackage{caption}
\usepackage{multirow}
\usepackage{array,booktabs,arydshln}
\usepackage{verbatim}
\usepackage[margin=0.75in]{geometry}
\def\BibTeX{{\rm B\kern-.05em{\sc i\kern-.025em b}\kern-.08em
    T\kern-.1667em\lower.7ex\hbox{E}\kern-.125emX}}
\begin{document}

\title{Classification by Re-generation-Towards Classification Based on Variational Inference
}

\author{\IEEEauthorblockN{Shideh Rezaeifar}
\IEEEauthorblockA{\textit{University of Geneva} \\
\textit{SIP group}\\
Geneva, Switzerland \\
shideh.rezaeifar@unige.ch}
\and
\IEEEauthorblockN{Olga Taran}
\IEEEauthorblockA{\textit{University of Geneva} \\
	\textit{SIP group}\\
	Geneva, Switzerland \\
	olga.taran@unige.ch}
\and
\IEEEauthorblockN{Slava Voloshynovskiy}
\IEEEauthorblockA{\textit{University of Geneva} \\
	\textit{SIP group}\\
	Geneva, Switzerland \\
  svolos@unige.ch}
\thanks{This research was partially supported by the SNF project 200021- 165672.}
}

\maketitle

\begin{abstract}
As Deep Neural Networks (DNNs) are considered the state-of-the-art in many classification tasks, the question of their semantic generalizations has been raised. To address semantic interpretability of learned features, we introduce a novel idea of classification by re-generation based on variational autoencoder (VAE) in which a separate encoder-decoder pair of VAE is trained for each class. Moreover, the proposed architecture overcomes the scalability issue in current DNN networks as there is no need to re-train the whole network with the addition of new classes and it can be done for each class separately. We also introduce a criterion based on Kullback-Leibler divergence to reject doubtful examples. This rejection criterion should improve the trust in the obtained results and can be further exploited to reject adversarial examples.

\end{abstract}


\section{Introduction}
\par Deep Neural Networks (DNNs) have achieved the state-of-the-art performances in many machine learning tasks. Nevertheless, recent advancements of Deep Neural Networks in image classification with near-human eye level of accuracy raised a variety of questions. Do DNNs develop an understanding of objects based on the training data and recognize them semantically? Or are they only very good mappers from the input to the label data? How would DNNs classify not-seen or unrecognizable objects?
\par Despite the high performance of DNNs, there is a doubt regarding their semantic generalization. Many researches have been conducted to validate or reject the hypothesis that DNNs develops an understanding of objects based on training data. In \cite{DNNLimitaions2017}, authors trained different architectures with regular images and tested their performances against negative examples. As negative examples have the same structure as the regular ones, humans can easily recognize them. However, in their experiments, the performances of DNNs dropped significantly when tested on negative images and they concluded that current methods in training DNNs fail to semantically recognize objects. 
\par Moreover, in the recent DNN architectures, an end-to-end training is usually applied where the entire architecture is optimized according to a specific loss function. This would cause a scalability problem as it is needed to re-train the entire network with the addition of new classes. In the real-world application of machine learning, one might need to add new classes on a regular basis, and as a result of current end-to-end training approach, the network should be re-trained.

\par Furthermore, in the classical machine learning scheme, it is assumed that the probe always belongs to one class. As a result, any not-seen object would be classified as one of the classes. This formulation leads to many issues in practice. As an example, in self-driving cars, the system might face classifying a road sign, which hasn't been seen before and that would probably cause a dangerous situation. One might claim that this issue would be resolved simply by adding not-seen examples in the training data. However, in a real-world application, it is infeasible to add all the not-seen examples to the training data. Moreover, training DNNs on both correct and incorrect classes does not bring a reasonable enhancement \cite{fooling_DNN}.

\par In this paper, we aim at testing the hypothesis that whether a good compressor or generator trained per class can be also a good classifier. This would lead to an idea of meaningful semantic training, which is not a case for the existing DNN architectures so far. Therefore, we introduce a concept of "classification by re-generation". For this purpose, we train Variational Auto-Encoder (VAE) \cite{VAE_Kingma2014} for each class of data and classify the probe based on the reconstructed output images. Moreover, based on this pipeline, there is no need to re-train the whole network with the addition of new classes and it can be done easily on the fly.
\par Furthermore, in our pipeline, we exploited a classification metric based on Kullback-Leibler divergence, which gives us a possibility to reject the doubtful cases. This rejection option might be of importance in many physical or medical experiments, where the trust in the obtained results is crucial. Additionally, such a metric is based on a complete distribution whereas the output of classification based on soft-max represents just a point-wise estimation. Last but not least, the ability of the network to reject is useful in making the system robust against adversarial examples. Thus, the main motivations behind the rejection option are threefold:
\begin{itemize}
	\item reliable rejection of doubtful examples (automatically without human intervention);
	\item trust in obtained results;
	\item robustness to semantic adversarial examples and unseen objects.	
\end{itemize}
\par In this study, we do not compete for an improvement in the classification accuracy with respect to the state-of-the-art end-to-end trained classification. Instead, we aim at introducing a new principle of classification based on re-generation towards scalability, interpretability, rejection, and trust in results.

\subsection{Related Work}
\par There have been enormous studies regarding the use of generative models for classification. In 2014, Kingma  et al. proposed a model for semi-supervised classification based on VAE \cite{VAE_semisup_kingma2014}. Their proposal achieved good performance, but it lacks the interpretability and acts as a black-box discriminator. 
\par Along the same line of research, in \cite{NIPS2017}, the authors proposed a framework to learn disentangled representations of data in the context of VAE. Their experiments showed promising results at the classification task. In \cite{Semi_sup_bayesian}, Gordon et al. explored the work detailed by Kingma et al. \cite{VAE_semisup_kingma2014} and introduced a slightly different inference network structure. The authors in \cite{Semi_sup_bayesian} used a bayesian neural network for label prediction. 
\par Despite their good performance in the context of classification based on VAE, previous works lack the interpretability of trained features. Moreover, due to an end-to-end training process, with the addition of new classes, the whole network should be retrained. The system scalability is of importance in the large-scale dataset and real-world applications. Furthermore, in our proposed model, we exploit the primary goal of VAE that is to generate for the purpose of classification. 
\par The rest of this paper is organized as follows. In section II, we briefly introduce the Variational Autoencoder framework. We then present our proposed model in section III. The experimental results are reported in section IV. Finally, section V concludes the paper.
\section{Background}

Generative models aim at learning the true distribution of data, $P(\mathbf{x})$, in order to generate new samples. To do so, they attempt to model the complex data by using latent variables. Two of the most commonly used and efficient approaches are  VAE \cite{VAE_Kingma2014} and Generative Adversarial Networks (GAN) \cite{GAN2014}. 
\par VAE was first introduced by  Kingma \& Welling in 2014 \cite{VAE_Kingma2014}. The model consists of two networks: Encoder and Decoder. The input data $\mathbf{x}$ is encoded to a latent representation $\mathbf{z}$ and then the samples $\mathbf{\hat{x}}$ are generated by a decoder from the latent space.  
\par Assume we are given a dataset, $\mathbf{X}=\{\mathbf{x}^1, . . . , \mathbf{x}^N\},$ consisting of N i.i.d. samples. In an unsupervised learning scheme, the log-likelihood of observations is maximized under a probabilistic mode  $P_{\boldsymbol\theta}(\mathbf{x})$ :
\begin{equation}
\log P_{\boldsymbol\theta}(\mathbf{X})=\sum_{i=1}^{N}{\log P_{\boldsymbol\theta}(\mathbf{x}^{i})}.
\end{equation}
\par In VAE, one assumes that the data were generated from low dimensional latent variables $\mathbf{Z}$. The probability distribution of latent variables is denoted by $P_{\boldsymbol\theta}(\mathbf{z})$ and the marginal likelihood  $P_{\boldsymbol\theta}(\mathbf{x})$ can be written as:
\begin{equation}
 P_{\boldsymbol\theta}(\mathbf{x})= \int P_{\boldsymbol\theta}(\mathbf{z}) P_{\boldsymbol\theta}(\mathbf{x}|\mathbf{z}) d\mathbf{z}.
\end{equation}
\par Due to the difficulty of working directly with marginal likelihood, a parametric inference model $Q_\psi(\mathbf{z} | \mathbf{x})$ is used. Thus, the marginal likelihood can be formulated as \cite{VAE_Kingma2014}:
\begin{equation}
\log P_{\boldsymbol\theta}(\mathbf{x})= D_{KL} ( Q_\psi(\mathbf{z} | \mathbf{x})  | P_{\boldsymbol\theta}(\mathbf{z} |\mathbf{x}) ) + \mathcal{L}(\mathbf{x};\boldsymbol\theta, \boldsymbol\psi)
\end{equation}
where $\boldsymbol\theta$ and $\boldsymbol\psi$ indicate the generative and variational parameters and $D_{KL}(.||.)$ is the Kullback-Leibler divergence. As the Kullback-Leibler divergence is non-negative, the term $\mathcal{L}(\mathbf{x};\boldsymbol\theta, \boldsymbol\psi)$ is considered to be a lower bound on the marginal likelihood. Therefore:
\begin{equation}
\log P_{\boldsymbol\theta}(\mathbf{x}) \geq  \mathcal{L}(\mathbf{x};\boldsymbol\theta, \boldsymbol\psi), 
\end{equation}
where
\begin{equation*}
\mathcal{L}(\mathbf{x};\boldsymbol\theta, \boldsymbol\psi) = E_{Q_{\boldsymbol\psi}(\mathbf{z} | \mathbf{x})}[\log P_{\boldsymbol\theta}(\mathbf{x} | \mathbf{z})] - D_{KL}(Q_{\boldsymbol\psi}(\mathbf{z} | \mathbf{x})||P_{\boldsymbol\theta}(\mathbf{z})).
\end{equation*}

\par The first term of the loss function corresponds to the reconstruction error of the decoder $P_{\boldsymbol\theta}(\mathbf{x} | \mathbf{z})$ and the second term is the Kullback-Leibler divergence between the prior distribution $P_{\boldsymbol\theta}(\mathbf{z})$ and the learned latent posterior $Q_{\boldsymbol\psi}(\mathbf{z}|\mathbf{x})$. The prior distribution is usually chosen to be a centered isotropic multivariate Gaussian $\mathbf{Z} \sim\mathcal{N}(\mathbf{0},\mathbf{I})$. Using the re-parameterization trick, VAE optimizes the lower bound \cite{VAE_Kingma2014}, \cite{rezende14}.

\section{Proposed Model}
\begin{figure}[t]
	\centering
	\includegraphics[width=7cm, height=5cm]{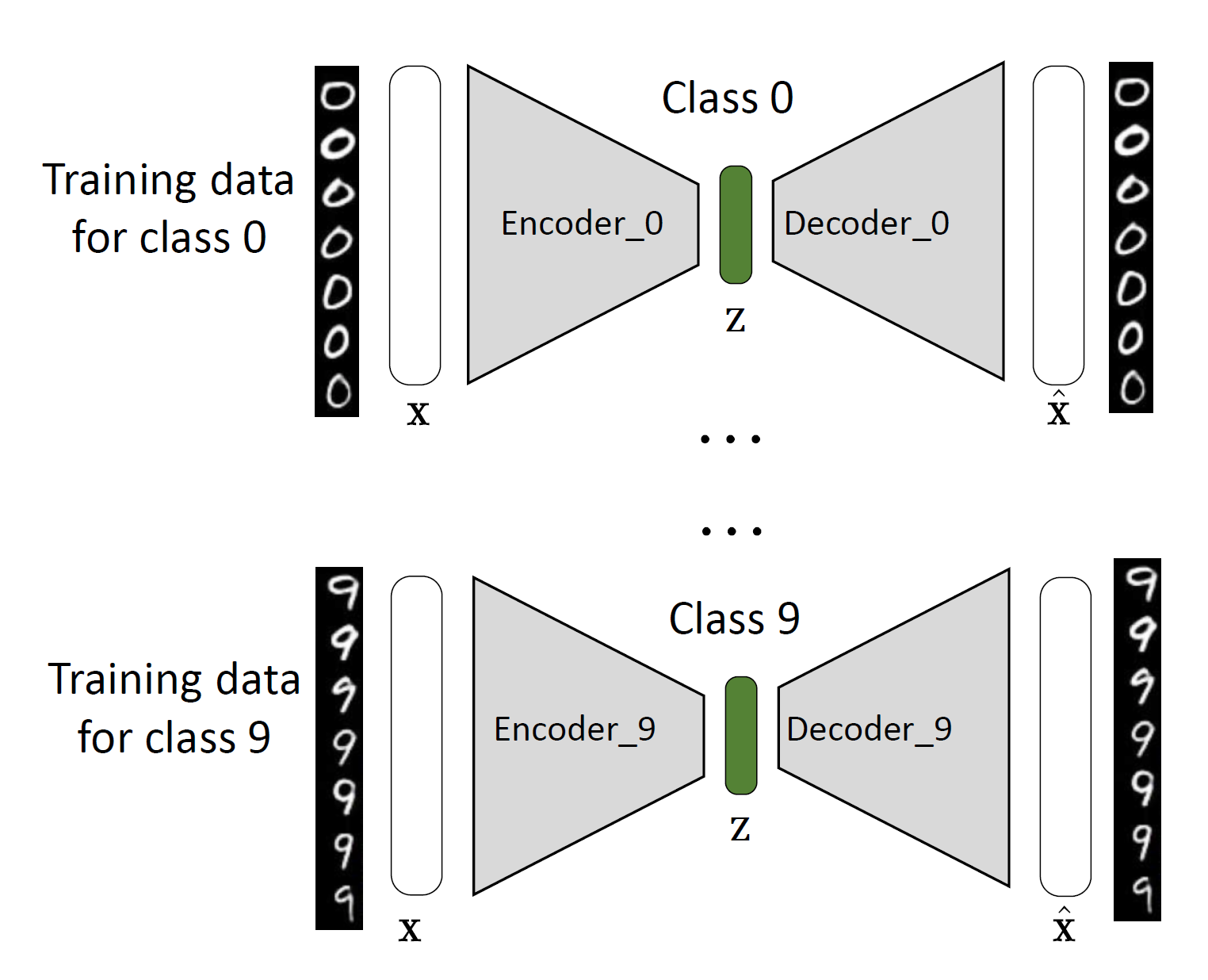}
	\caption{Training VAE for each class of data.}
	\label{fig:train_VAE}
\end{figure}
\begin{figure}[t]
	\begin{subfigure}{3.85cm}
	\centering
	\includegraphics[width=3.85cm, height=5cm]{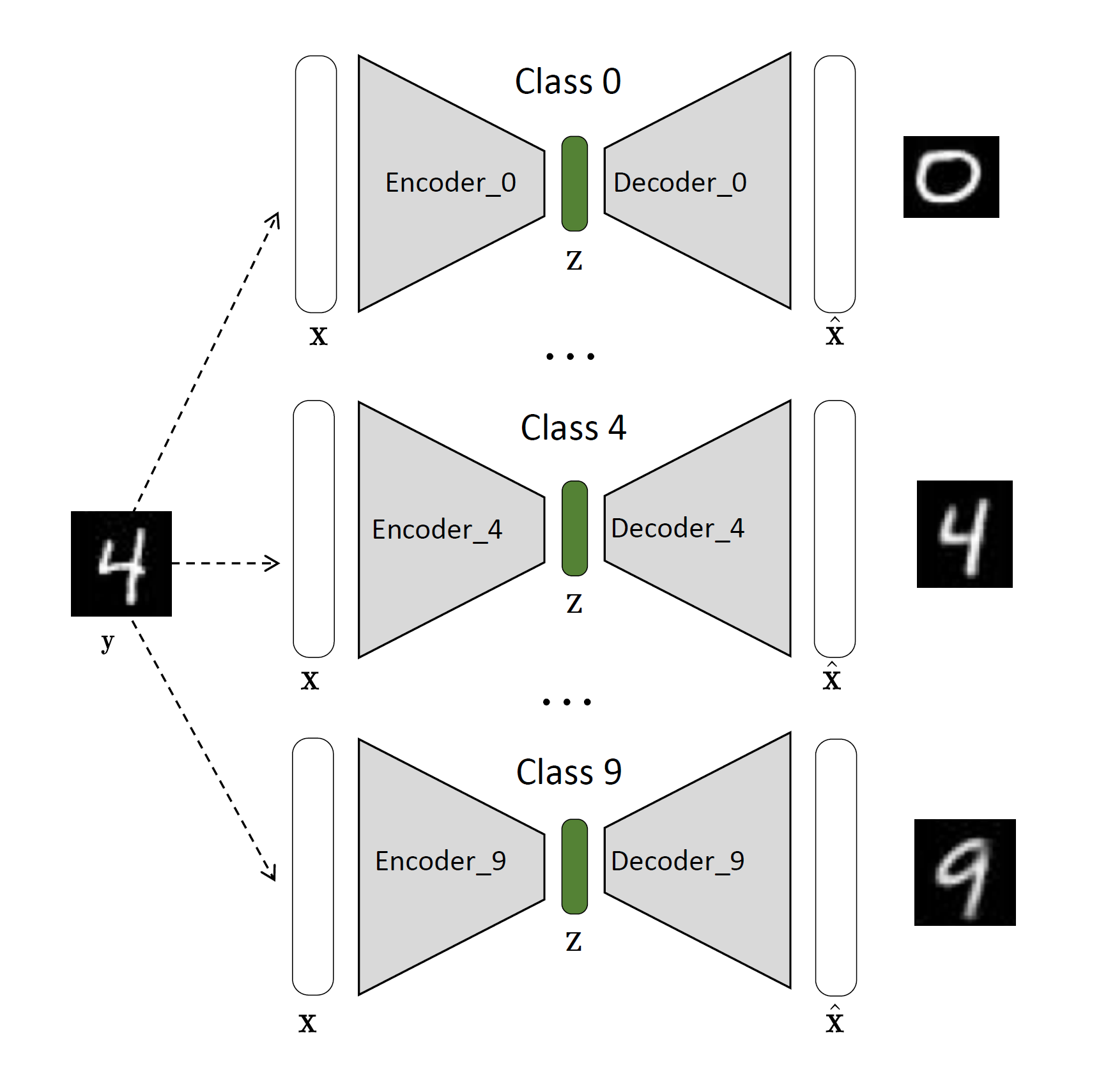}
    \end{subfigure}
	\begin{subfigure}{5cm}
	\centering
	\includegraphics[width=5cm, height=5cm]{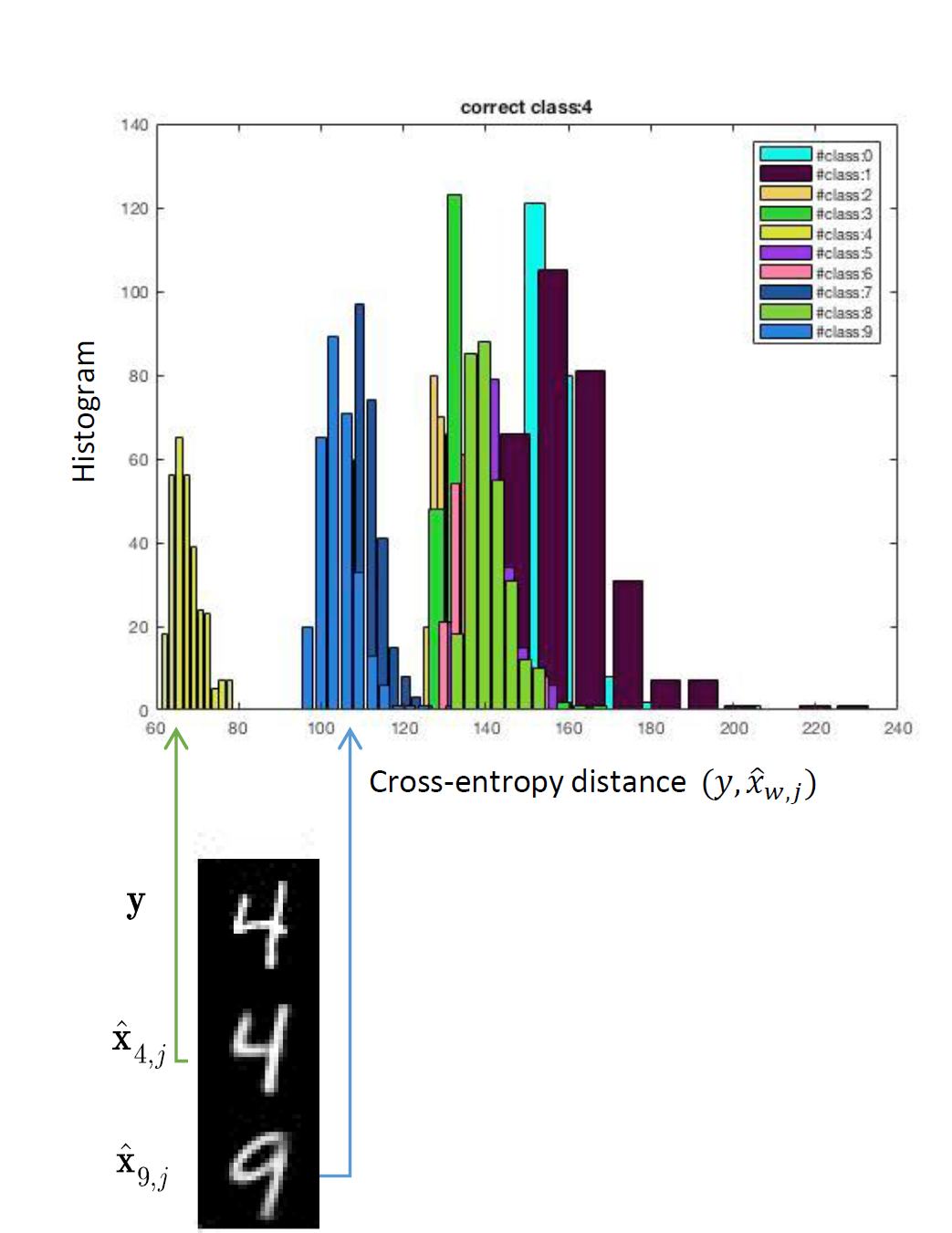}
\end{subfigure}
	\caption{Recognition: we generate multiple outputs using the stochastic behavior of VAE.}
	\label{fig:recognition_VAE}
\end{figure}
VAE has been already proposed in semi-supervised classification settings \cite{VAE_Kingma2014}. In this paper, we investigate the potential of VAE from another perspective. The main idea is to exploit the primary objective of VAE that is the generation for the purpose of classification, whereas the VAE for each class is trained in an unsupervised way.
\par The main principle is to train its own VAE for each class in a way to capture the statistical distribution of that class as shown in Fig. \ref{fig:train_VAE}. Given a probe image, we pass it through each encoder-decoder pair of VAE and we argue that the best reconstruction would be achieved, if the probe belongs to that class. An example of recognition is shown in Fig. \ref{fig:recognition_VAE}. The main argument behind this architecture is the interpretability of learned features and scalability.

\par Variety of metrics can be applied to measure the similarity of input and reconstructed outputs. In this study, we focus on cross-entropy as a measure of similarity. Due to a hidden random state in VAE, point-wise estimation based cross-entropy has a lot of drawbacks. To address this issue, we re-generate the output 300 times for the same probe and make the decision based on an estimate of PDF of cross-entropies. This can be further extended to any other cost functions. Moreover, the PDF estimate of  cross-entropies enables us to define a criterion for rejecting doubtful examples based on Kullback-Leibler divergence. 
\par
\begin{figure}[b]
	\centering
	\includegraphics[width=7.5cm, height=3.5cm]{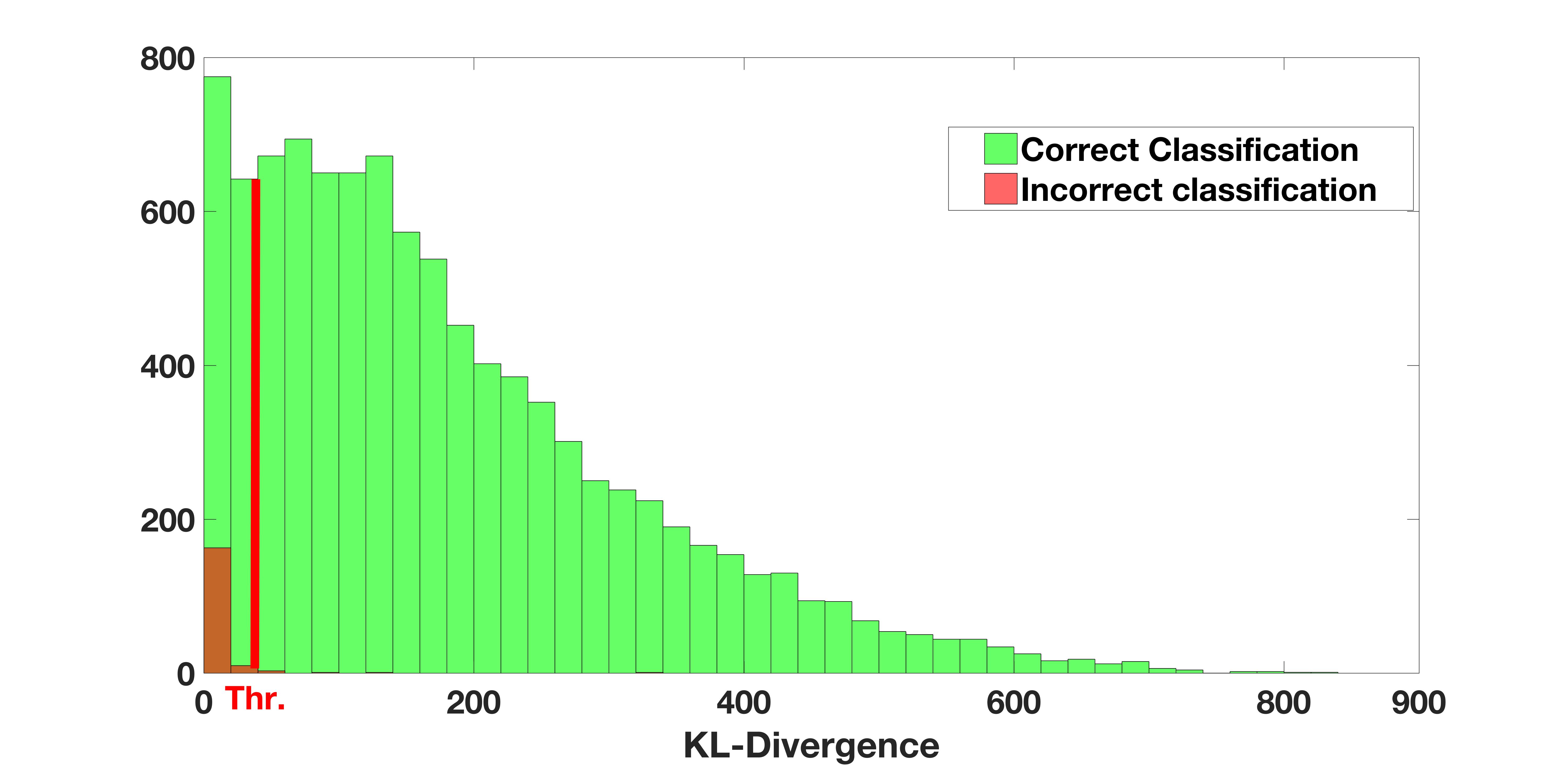}
	\caption{The distributions of Kullback-Leibler divergence for correct and incorrect classifications.}
	\label{fig:KL_hist}
\end{figure}

 The rejection criterion is defined as the Kullback-Leibler divergence between the PDF of the second highest score and the PDF of the highest one. The highest score is considered as those that have the smallest reconstruction distortion. If this divergence is below a certain threshold, the classifier would reject it as a doubtful probe.  Kullback-Leibler divergence distributions for correct and incorrect classifications for MNIST dataset are shown in Fig.\ref{fig:KL_hist} . As one can see, we would reject the doubtful probes for which their divergences are below the threshold $Thr$ at the cost of missing some correct classifications reported as $P_{miss}$ in section IV.A. 
ُ\par
\begin{figure}[t]
	\begin{subfigure}{6cm}
		\centering
		\includegraphics[width=6cm, height=3.5cm]{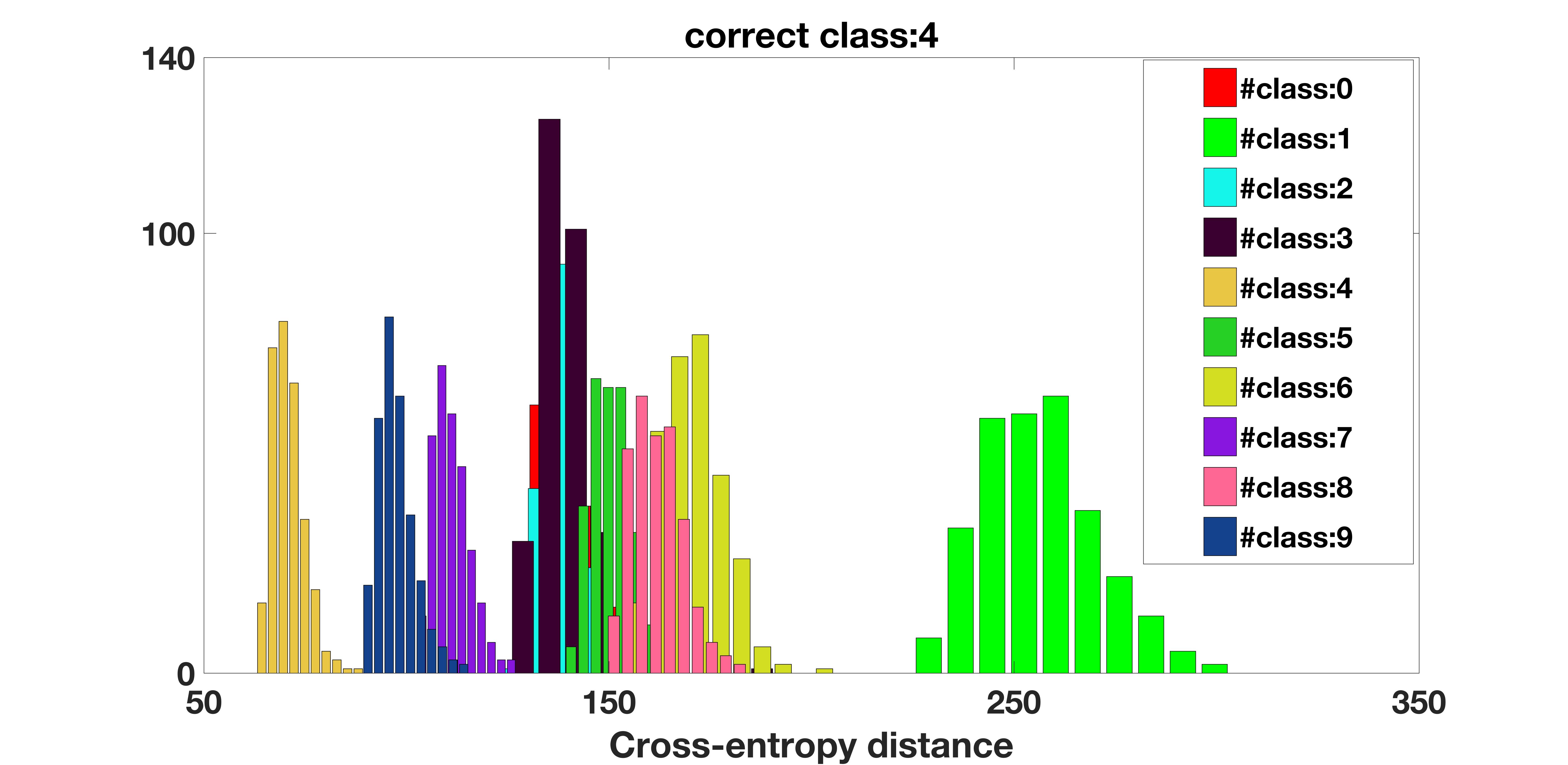}
		\caption{Correctly classified example}
	\end{subfigure}
	\begin{subfigure}{2.5cm}
		\centering
		\includegraphics[width=2.5cm, height=3.5cm]{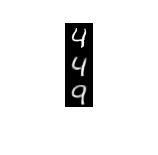}
	\end{subfigure}
	
	\begin{subfigure}{6cm}
		\centering
		\includegraphics[width=6cm, height=3.5cm]{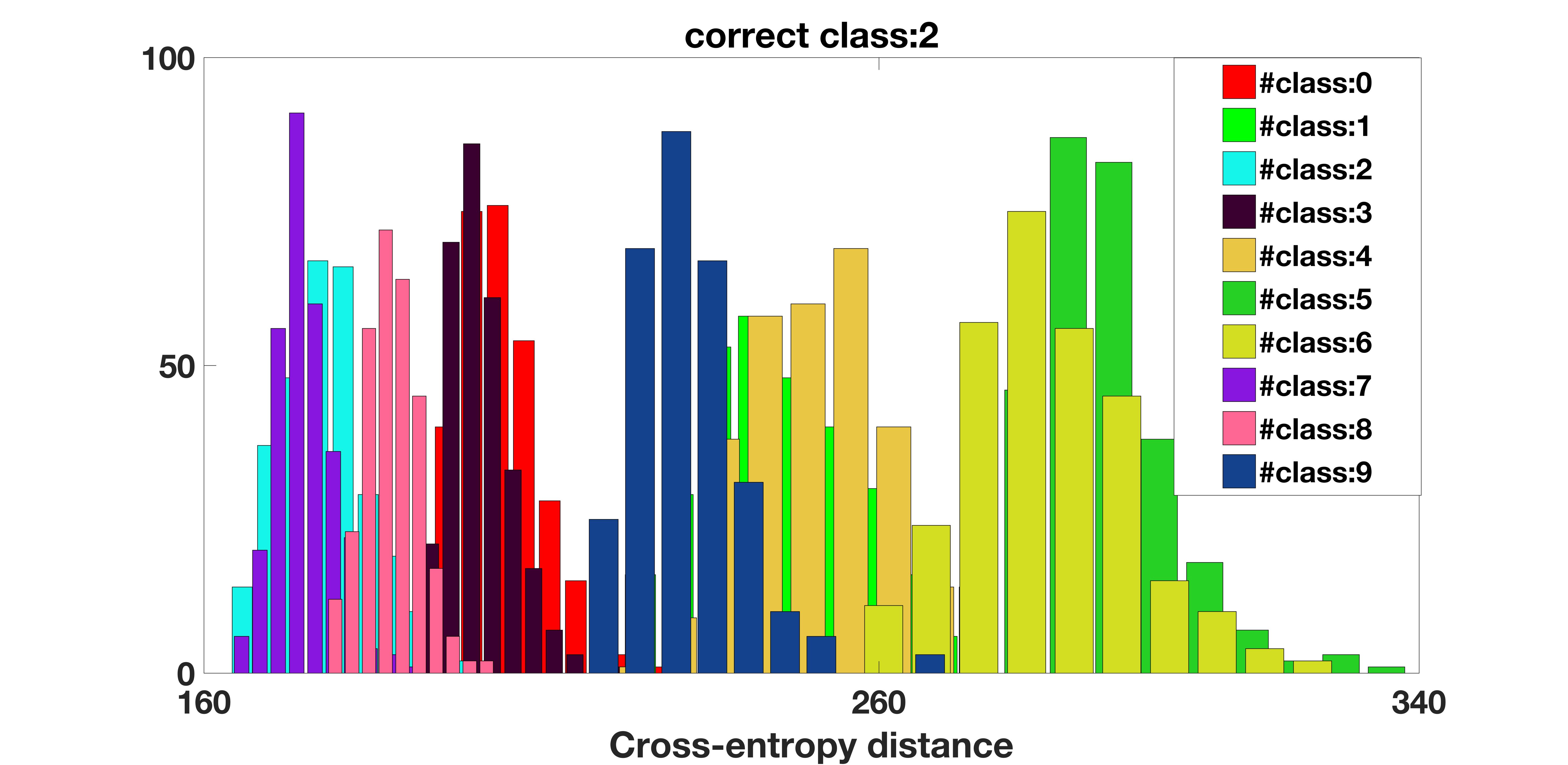}
		\caption{Rejected example}
	\end{subfigure}
	\begin{subfigure}{2.5cm}
		\centering
		\includegraphics[width=2.5cm, height=3.5cm]{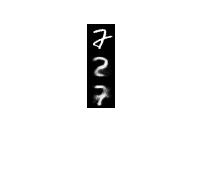}
		
	\end{subfigure}
	\caption{Examples of a) correctly recognized b) rejected probes.}
	\label{Fig:examples_correct_reject}
	
\end{figure}

 To better clarify the need for rejection criterion, examples of correct classification and probable misclassification for MNIST dataset are shown in Fig. \ref{Fig:examples_correct_reject}. In this figure, the histograms of cross-entropy distances as well as corresponding images of input, highest score and second highest score are shown. In the case of correct classification, the histogram of the correct class is well-separated from others. However, in the second case, there is an overlap between the histograms of classes. More importantly, the correct class "2" is not recognizable even by humans. 
\par As shown in Fig. \ref{Fig:examples_correct_reject}, the classes are clearly separated in the correct case and overlap in the probably incorrect case.

\par Additionally, this pipeline can be further extended for large-scale datasets. In the current architecture, for a dataset of $M$ classes, $M$ pairs of encoder-decoder are required. However, assuming a tree-based structure, this number can be reduced to $\log_2{M}$. In
the case of tree-based structure, the classes are divided into the groups of varying size $(M/2,..,3,2,1)$ based on a similarity measurement in a form of a tree. Therefore, instead of training VAE for each class, the VAEs in each layer of the tree are trained on a group of classes with varying size. To classify a probe image, in each layer the distances between the input and reconstructed outputs of that layer are obtained. Afterward, given this distance, the decision of the next group to test in the next layer is made. This process is repeated until the probe image is finally classified in the last layer.
\section{Experimental results}
\par For our experiments, MNIST  and Fashion-MNIST datasets were used as they are commonly known and tested for classification and generation. We have used TensorFlow \cite{tensorflow} to implement the standard VAE with two-layer MLPs of 500 hidden units as encoder and decoder models. The default dimensionality of latent variables we set to 15. For learning, we used Adam \cite{adam} with a learning rate set to 0.001. 
\par The purpose of these experiments is to validate or reject the hypothesis that whether a well-trained generator can be used as a good classifier. We do not aim at competing with the state-of-the-art end-to-end trained classifiers, instead, we want to evaluate a new principle of classification for better scalability and interpretability. 
\par In all the following experiments, probability of miss $P_{miss}$ and accuracy $Acc.$ are defined as:

\begin{equation}
\begin{aligned}
P_{miss}&=\frac{\text{Number of correct and rejected}}{\text{Total number of test data}} \\
Acc.&=\frac{\text{Number of correct and not rejected}}{\text{Total number of not rejected data}}
\end{aligned}	
\end{equation}
Additionally, for the outputs of VAEs, their corresponding probability distributions of distances were obtained and ranked based on the minimum distance. Hence, the rejection criterion is defined as:
\begin{equation}
\begin{cases}
\text{Classify }, & \text{if} ~~ D_{KL}(P_{first} || P_{second} ) \geq Thr.  \\
\text{Reject }, &\text{if} ~~ D_{KL}(P_{first} || P_{second} ) < Thr. 
\end{cases}
\end{equation}
where $P_{first}$ and $P_{second}$ are the probability distributions of the first and second class in the ranking, respectively.
\subsection{Impact of VAE parameters}
In this section, we investigate the impact of different parameters of VAE, namely, dimensionality of latent variables and variance of random state, on the classification accuracy for MNIST and Fashion-MNIST datasets.
\par As shown in Table \ref{Tab:diff_dim_mnist} for MNIST dataset, the accuracy of system increases as we increase the dimensionality of latent variables. However, at the dimensionality of 20, we noticed over-fitting and a drop in the performance of the system. Moreover, the results for Fashion-MNIST dataset are reported in Table \ref{Tab:diff_dim_fashion} and the best performance is achieved for the dimensionality of 20.
\par Moreover, we also investigated the effect of randomness in the decoder on the system accuracy, which are reported in Table \ref{Tab:diff_sigma_mnist} and \ref{Tab:diff_sigma_fashion}. The randomness in the decoder is defined by the variance of the noise in the re-parametrization trick. We noticed that decreasing the variance $\sigma^2$, would actually improve the performance. 

\begin{table}[t]
	\centering
	\caption{$Acc$ and $P_{miss}$ for different dimensionalities of latent variables $\mathbf{z}$ on MNIST dataset.}
	\def\arraystretch{1.5}
	\begin{tabular}{|c|c|c|c|c|c|c|}
		\hline
		\multirow{ 2}{*}{\textbf{Thr.}}&  \multicolumn{2}{|c|}{\textbf{$dim_z=20$}} & \multicolumn{2}{|c|}{\textbf{$dim_z=15$}} & \multicolumn{2}{|c|}{\textbf{$dim_z=10$}}  \\ \cline{2-7}
		& Acc. & $P_{miss}$ &Acc. & $P_{miss}$ & Acc. & $P_{miss}$
		\\\specialrule{3pt}{0pt}{0pt}
		1 & 98.82 &0.012& \textbf{98.96}& 0.009 &98.41&0.048 \\ \hline
		5 & 99.41 &0.039& \textbf{99.50} &0.029 &98.99&0.07\\ \hline
		10& 99.69 &0.07& \textbf{99.73} &0.048&99.24&0.087 \\ \hline
		15& 99.81 &0.01& \textbf{99.82} &0.065&99.35&0.1\\ \hline
		20& 99.85 &0.13& \textbf{99.86} &0.078&99.42&0.11\\ \hline
		no rejection &98.07&-&\textbf{98.27} &-&98.04&-\\ \hline
		
	\end{tabular}	
	\label{Tab:diff_dim_mnist}
\end{table}
\begin{table}[t]
	\centering
	\caption{Impact of different values of $\sigma^2$ on MNIST dataset}
	\resizebox{\columnwidth}{!}{%
		\def\arraystretch{1.5}
		\begin{tabular}{|c|c|c|c|c|c|c|c|c|c|}
			\hline
			&\multicolumn{3}{|c|}{\textbf{$dim_z=10$}} & \multicolumn{3}{|c|}{\textbf{$dim_z=15$}} & \multicolumn{3}{|c|}{\textbf{$dim_z=20$}}  \\ \cline{1-10}
			$\sigma^2$&$.5$&$.75$&$1.25$&$.5$&$.75$&$1.25$&$5$&$.75$&$1.25$
			\\\specialrule{3pt}{0pt}{0pt}
			Acc.&\textbf{98.36}&98.27&98.17&98.33&98.22&98.14&98.10&98&97.93\\ \hline	
		\end{tabular}%
	}
	\label{Tab:diff_sigma_mnist}	
\end{table}

\begin{table}[b]
	\centering
	\caption{$Acc$ and $P_{miss}$ for different dimensionalities of latent variables $\mathbf{z}$ on Fashion-MNIST dataset}
	\def\arraystretch{1.5}
	\begin{tabular}{|c|c|c|c|c|c|c|}
		\hline
		\multirow{ 2}{*}{\textbf{Thr.}}&  \multicolumn{2}{|c|}{\textbf{$dim_z=25$}} & \multicolumn{2}{|c|}{\textbf{$dim_z=20$}} & \multicolumn{2}{|c|}{\textbf{$dim_z=15$}}  \\ \cline{2-7}
		& Acc. & $P_{miss}$ &Acc. & $P_{miss}$ & Acc. & $P_{miss}$
		\\\specialrule{3pt}{0pt}{0pt}
		1 & 92.23 &0.07& \textbf{93.13}& 0.068 &82.89&0.061 \\ \hline
		2 & 94.34 &0.10& \textbf{94.96} &0.11 &84.35&0.099\\ \hline
		3& 95.88 &0.13& \textbf{96.15} &0.14&85.19&0.12 \\ \hline
		4& 96.88 &0.16& \textbf{97.02} &0.16&85.71&0.14\\ \hline
		5& 97.5 &0.18& \textbf{97.6} &0.18&86.03&0.16\\ \hline
		no rejection &88.61&-&\textbf{89} &-&79.66&-\\ \hline
		
	\end{tabular}	
	\label{Tab:diff_dim_fashion}
\end{table}

\begin{table}[b]
	\centering
	\caption{Impact of different values of $\sigma^2$  Fashion-MNIST}
	\resizebox{\columnwidth}{!}{%
		\def\arraystretch{1.5}
		\begin{tabular}{|c|c|c|c|c|c|c|c|c|c|}
			\hline
			&\multicolumn{3}{|c|}{\textbf{$dim_z=10$}} & \multicolumn{3}{|c|}{\textbf{$dim_z=15$}} & \multicolumn{3}{|c|}{\textbf{$dim_z=20$}}  \\ \cline{1-10}
			$\sigma^2$&$.5$&$.75$&$1.25$&$.5$&$.75$&$1.25$&$5$&$.75$&$1.25$
			\\\specialrule{3pt}{0pt}{0pt}
			Acc.&88.7&88.56&88.83&\textbf{89.03}&89&88.84&79.36&80.96&88.33\\ \hline	
		\end{tabular}%
	}
	\label{Tab:diff_sigma_fashion}	
\end{table}

\subsection{Impact of limited label data}
Furthermore, several experiments were conducted to test the model performance in a semi-supervised setting. In these experiments, a limited number of labeled training samples, N, was used to train the network on the MNIST dataset. 

There are various approaches for the semi-supervised classification including Transductive SVMs (TSVM) \cite{TSVM} as an extension of SVM for limited labeled data. In the \cite{Tangent}, the authors proposed two approaches, namely, Contrastive Auto-Encoders (CAE) and Manifold Tangent Classifier (MTC), based on neural networks to achieve high performance for semi-supervised classification. In CAE, the authors trained a two-layer deep network with CAE objective function, whereas MTC is trained with tangent propagation.
\par In Table \ref{Tab.semi_supervised}, we compared our result with the state-of-the-art in semi-supervised classification setting. Although the performance of the proposed model is not better than the state-of-the-art, it is still competitive. \par Considering the fact that the implementation was based on vanilla VAE without any pre-processing or techniques such as normalizing flows, the obtained results validate our hypothesis that classification based on re-generation and more specifically VAE has the potential to be further investigated.

\begin{table}[t]
	\centering
	\caption{ Semi-supervised classification error based on VAE on MNIST dataset.}
	\def\arraystretch{1.75}
	\begin{tabular}{|c|c c c c| }
		\hline
	    Method & N=600 &N=1000&N=3000& Supervised\\
	    \hline \hline
	    NN &11.44&10.7&6.04&- \\
	   CNN &7.57&6.45&3.35&-\\
	   TSVM\cite{TSVM} &6.16&5.38&3.45&-\\
	    MTC \cite{Tangent} &5.13&3.64&2.57&-\\
	    M1+M2 \cite{VAE_semisup_kingma2014}&4.94&3.6&3.92&.96\\
	    Disentangled   &3.84&2.88&1.57&-\\
	    Proposed Model&9.77&7.64&4.83 &1.73\\
	    \hline
	
	\end{tabular}
	\label{Tab.semi_supervised}
\end{table}

\subsection{Rejecting the not-seen objects}
\par In order to evaluate our rejection criterion, we designed an experiment in which we trained our network on MNIST dataset and tested on the Fashion-MNIST dataset. The percentages of rejection for different values of threshold as well as $P_{miss}$ are reported in Table \ref{Tab.rejection}. The accuracy of the original dataset MNIST for the same threshold can be found in Table \ref{Tab:diff_dim_mnist}.
\begin{table}[t]
	\centering
	\caption{ Percentages of rejection and $P_{miss}$. }
	
	\def\arraystretch{1.5}
	\begin{tabular}{|c| c c |}
		\hline
		Threshold & Percentage of rejection & $P_{miss}$  \\ \hline \hline
		10 &92.65&0.048 \\
		15 &95.1& 0.063\\
		20 &96.4&0.077 \\
		30&97.76&0.1093 \\
		\hline
		
	\end{tabular}%
	
	\label{Tab.rejection}
\end{table}
\subsection{Interpretability of learned features}
\begin{figure}[b]
	\begin{subfigure}{4.25cm}
		\centering
		\includegraphics[width=3.5cm, height=2cm]{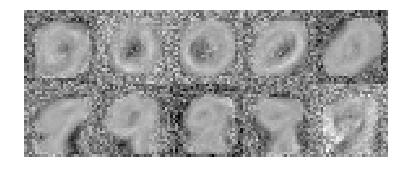}
		\caption{VAE for class "0" and "9"}
	\end{subfigure}
	\begin{subfigure}{4.25cm}
		\centering
		\includegraphics[width=3.5cm, height=2cm]{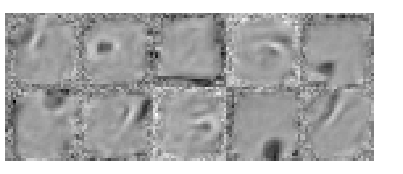}
		\caption{common VAE}
	\end{subfigure}

	\caption{The visualizations of filters in the last layer of decoder.}
	\label{fig:filter_vis}
\end{figure}

In order to investigate the interpretability of the learned features, we visualized the filters in the last layer of the decoder for different classes and compared them with those of one common VAE trained on all of classes. As shown in Fig.\ref{fig:filter_vis}, in the case of class specific VAE, the filters obviously follow the structure and shapes of the corresponding class, whereas in the common VAE, no specific pattern or structure is observed.

\section{Conclusions}
\par Although deep neural networks have shown a great performance in a variety of machine learning tasks, they do not semantically generalize well \cite{DNNLimitaions2017}. Attempting to address this problem, we proposed the idea of "classification by re-generation" . In our proposed architecture, for each class of data, an encoder-decoder pair of VAE is trained and the classification decision of the probe image is based on the reconstructed outputs of each class.  Moreover, this architecture  resolves the scalability issue of existing end-to-end trained classifiers, as it doesn't need to re-train the whole network with the addition of new classes. 
\par Furthermore, the classification decision is based on a complete PDF of cross-entropy distances. Given the PDF of distances, we introduced a rejection criterion to avoid doubtful probes. In this way, we can ensure a certain level of trust in the produced result that can be semantically and visually validated. 
\par The experimental results validated our idea of classification by re-generation and demonstrated that the proposed model has the potential for further investigation.

\section{Future work}
Future work includes implementing the tree-based structure to avoid the scalability issues in large-scale datasets. In addition to that, we intend to apply our approach to other datasets such as CIFAR-10 and boost our encoders and decoders with more recent techniques such as normalizing flows. We will also look into a successive VAE based on residuals to overcome the current problem of VAE related to the blurred nature of generated images.

\bibliographystyle{IEEEbib}
\bibliography{EUSIPCOreference}


\end{document}